\documentclass[10pt, conference]{IEEEtran}
 \IEEEoverridecommandlockouts
\usepackage{cite}
\usepackage{amsmath,amssymb,amsfonts}
\usepackage{algorithmic}
\usepackage{graphicx}
\usepackage{textcomp}
\usepackage{xcolor}
\usepackage{booktabs}
\usepackage{subcaption}
\usepackage{multirow}
\usepackage{url}
\usepackage{float}
\usepackage{xcolor}
\def\BibTeX{{\rm B\kern-.05em{\sc i\kern-.025em b}\kern-.08em
    T\kern-.1667em\lower.7ex\hbox{E}\kern-.125emX}}

\usepackage{newtxtext,newtxmath}

\DeclareGraphicsExtensions{.pdf,.png,.jpg}

\begin{document}

% \title{Automatic Algorithm Selection for Combinatorial Optimization Tasks in Resource-Constrained Environments}
\title{SLA-Centric Automated Algorithm Selection Framework for Cloud Environments}

\author{\IEEEauthorblockN{Siana Rizwan}
\IEEEauthorblockA{\textit{School of Computing} \\
\textit{Queen's University}\\
Kingston, Ontario, Canada \\
siana.rizwan@queensu.ca}
\and
\IEEEauthorblockN{Tasnim Ahmed}
\IEEEauthorblockA{\textit{School of Computing} \\
\textit{Queen's University}\\
Kingston, Ontario, Canada \\
tasnim.ahmed@queensu.ca}
\and
\IEEEauthorblockN{Salimur Choudhury}
\IEEEauthorblockA{\textit{School of Computing} \\
\textit{Queen's University}\\
Kingston, Ontario, Canada \\
s.choudhury@queensu.ca}
}

\maketitle

\begin{abstract}

Cloud computing offers on-demand resource access, regulated by  Service-Level Agreements (SLAs) between consumers and Cloud Service Providers (CSPs). SLA violations can impact efficiency and CSP profitability. In this work, we propose an SLA-aware automated algorithm-selection framework for combinatorial optimization problems in resource-constrained cloud environments. The framework uses an ensemble of machine learning models to predict performance and rank algorithm-hardware pairs based on SLA constraints. We also apply our framework to the 0-1 knapsack problem. We curate a dataset comprising instance specific features along with memory usage, runtime, and optimality gap for 6 algorithms. As an empirical benchmark, we evaluate the framework on both classification and regression tasks. Our ablation study explores the impact of hyperparameters, learning approaches, and large language models' effectiveness in regression, and SHAP-based interpretability.
\end{abstract}

\begin{IEEEkeywords}
Cloud Computing, Service-level agreements, 0-1 knapsack, Machine learning, SHAP, Large Language Models 
\end{IEEEkeywords}

%3rd update
\section{Introduction}
Modern computing heavily depends on cloud environments, where computing resources are delivered as services under Service-Level Agreements (SLAs) by Cloud Service Providers (CSPs). SLAs define Quality of Service (QoS) parameters and associated costs. Fulfilling SLAs is crucial for maintaining service reliability, optimizing resource utilization, and avoiding financial losses or customer dissatisfaction.  In this work, we focus on selecting algorithms for combinatorial optimization problems within an SLA-based cloud environment. To the best of our knowledge, existing algorithm selection approaches are mostly static and independent of resource availability. Inefficient selection increases runtime and costs, demonstrating the need for intelligent, learning-based methods to predict performance and facilitate real-time resource management \cite{khan2022machine}.

\begin{figure}[htbp]
\centerline{\includegraphics[width=\columnwidth]{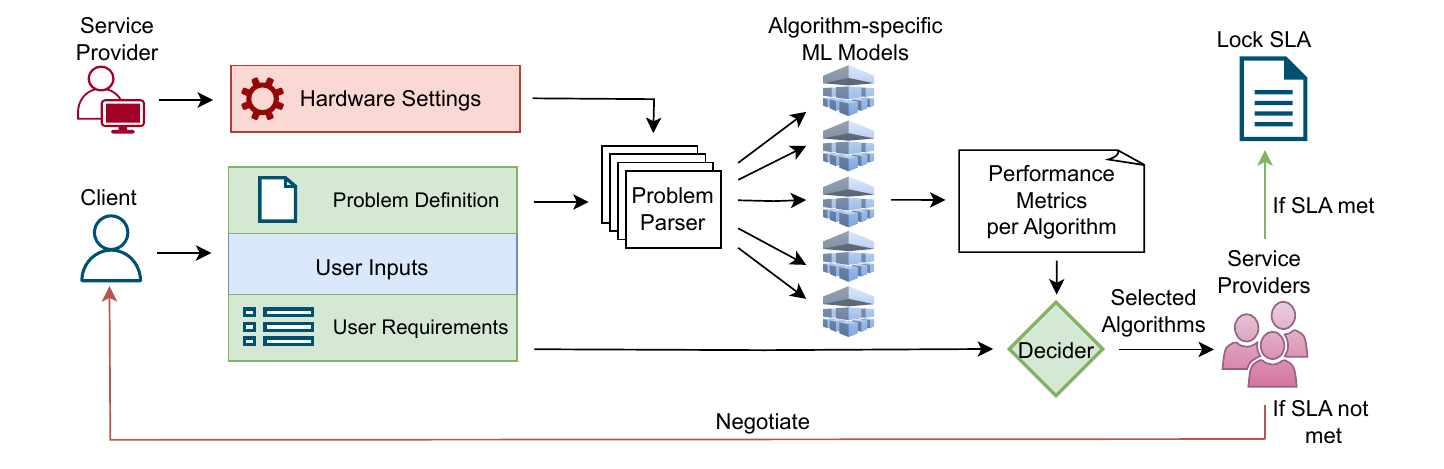}}
\caption{SLA-Centric algorithm selection framework}
\label{fig: overview_framework}
\end{figure}

To this end, we propose a Machine Learning (ML)-based automated algorithm selection framework for resource-constrained cloud environments. As shown in Figure \ref{fig: overview_framework}, the process starts with a problem parser that identifies the optimization problem type from the user-defined input. Subsequently, relevant problem instances and hardware constraints are forwarded to algorithm-specific ML models, which predict key performance metrics such as solution time, memory usage, and optimality gap for each instance-hardware pair. Here, the optimality gap refers to how close a solution is to the best possible (optimal) outcome. This metric is essential for CSPs to ensure solution quality in complex optimization tasks where near-optimal results might be preferred by the end-users (as the user might be aware of the hardness of the problem and not seek the optimal). These predictions are then evaluated by a `Decider' module against the user's SLA requirements to select the best-fitting algorithm. If the requirements are met, the SLA is finalized; otherwise, a negotiation is initiated based on the violated constraints.

 As a representative use case, we apply our framework to the 0-1 Knapsack Problem (KP), a classical NP-hard problem that is applicable in real-world scenarios, including logistics, supply chain management, resource scheduling, and planning \cite{huerta2020anytime}. Furthermore, KP is widely used in telecommunications and networking systems, as well. For instance, KP-based heuristics have been used for efficient edge server placement in 5G \cite{TIWARI2024222} and resource allocation in fog-based IoT systems, improving both energy usage and cost-effectiveness \cite{shruthi2022resource}. We thus consider both profit-maximization and profit-minimization variants of the problem to reflect diverse performance trade-offs. Since the framework is based on algorithm performance and instance-specific features, it is easily adaptable to other optimization tasks, given a comparable instance set with instance and hardware-specific features. Our key contributions are: (1) we propose a generalizable, SLA-driven algorithm selection framework for cloud environments to support resource-aware decision-making by CSPs; (2) we evaluate the framework on our curated dataset using both classification and regression-based ML models with hyperparameter tuning and ensemble techniques; (3) we also explore Q-learning and SARSA-based Reinforcement Learning (RL) as alternative predictors; (4) furthermore, we compare the performance of zero-shot Large Language Models (LLMs) with finetuned regressors.

\section{Literature Review}
\label{lit_review}
Recent studies have focused on ML-based algorithm selection for combinatorial problems. For Travelling Salesman Problem (TSP), Kerschke et al. \cite{kerschke2018leveraging} conducted a comprehensive comparative analysis involving five state-of-the-art heuristic algorithms (LKH, EAX, MAOS, and restart variants) over 1845 problem instances. They proposed an algorithm selection framework using classification, regression, and paired regression with models such as Random Forest (RF), Support Vector Machine (SVM), and Recursive Partitioning, and Multivariate Adaptive Regression Splines. Heins et al. \cite{heins2021potential} demonstrated that theoretically normalized features enhance SVM and RF performance for the same. For 0-1 KP, Jooken et al. \cite{jooken2022new} introduced hard instances challenging even advanced algorithms, which Huerta et al. \cite{huerta2020anytime} used to develop an anytime selection framework employing RF, Gradient Boosting, and MLP. Furthermore, Messelis et al. \cite{MESSELIS2014511} proposed an automatic algorithm selection approach based on empirical hardness models, where ML is used to predict algorithm performance from instance features for multi-mode resource-constrained project scheduling problem.
These works emphasize the value of learning-based approach for algorithm selection.
Additionally, Lagoudakis et al. \cite{lagoudakis2000reinforcement} applied RL by modeling algorithm selection as an MDP and used Q-learning to dynamically minimize solution time.\\
However, existing works on algorithm selection overlook hardware constraints, leading to potential SLA violations. To bridge this gap, our framework predicts performance metrics for algorithms with a novel focus on the optimality gap under varied settings. By integrating these predictions into the SLA finalization process, it supports an efficient, resource-aware algorithm selection for complex optimization tasks.

\section{Methodology}
We curate 200 instances for 0-1 KP based on Jooken et al. \cite{jooken2022new} and augment them with synthetic instances by introducing controlled noise to vary weight-profit correlation and instance hardness, while maintaining consistent item counts and capacities. Each instance is represented using 22 statistical and domain-specific features, following Huerta et al. \cite{huerta2020anytime}. The instances are solved using 6 algorithms/solvers: greedy, Dynamic Programming (DP), Genetic Algorithm (GA), Branch and Bound (BnB), Gurobi, and Google OR-Tools. A uniform time limit of 300 seconds is applied. For each algorithm-instance pair, we record solution time ($T_s$), memory usage ($M_s$), and optimality gap ($O_s$), using Gurobi’s output as the optimal reference. To replicate real-world cloud environments, we run all algorithms under varied hardware configurations (RAM: 4–256 GB, sampled by a factor of 2; CPU cores: 8 and 32), producing 2800 samples per algorithm.

For predictive modeling, we evaluate standalone classifiers and regressors per algorithm and KP variant: classifiers predict performance categories, while regressors estimate exact values. We test 7 ML models: Logistic/Linear Regression (LR), Decision Tree (DT), RF, MLP, SVM, CatBoost, and 1D CNN. These models were selected to evaluate a combination of traditional ML and deep learning methods, including the commonly used ones in \cite{heins2021potential, kerschke2018leveraging, huerta2020anytime}. However, we emphasized on traditional ML models more due to their interpretability and lower computational overhead, which are advantageous for deployment in resource-constrained environments. Final predictions are obtained via equal-weighted top-3 ensembles. Additionally, we implement RL-based frameworks (Q-learning, SARSA) for the profit-maximization variant, modelled as an MDP \cite{lagoudakis2000reinforcement}.

\section{Experimental Evaluation}
\label{sec_experimental_eval}
 The dataset is split into a 60--20--20 ratio for training, validation, and testing. Due to imbalanced class distribution, classification models are evaluated using both accuracy and F1-score. Regression and RL models are evaluated using Root Mean Squared Error (RMSE) and Coefficient of Determination ($R^2$). Results indicate that while regressors offer precise numeric predictions, ensemble-based classifiers provide more stable and interpretable outcomes.

\subsection{Classification-based Model Performance}
\begin{table}
\caption{Models performance for each algorithm (Classification $|$ Maximization)}
\centering
\scriptsize
\setlength{\tabcolsep}{2pt}{
\begin{tabular}{|c|l|cc|cc|cc|}
\hline
\textbf{Algorithm} & \textbf{Model} & \multicolumn{2}{c|}{\textbf{Solution Time}} & \multicolumn{2}{c|}{\textbf{Optimality Gap}} & \multicolumn{2}{c|}{\textbf{Peak Memory}} \\
\cline{3-8}
 & & \textbf{Accuracy} & \textbf{F1-Score} & \textbf{Accuracy} & \textbf{F1-Score} & \textbf{Accuracy} & \textbf{F1-Score} \\
\hline
\multirow{8}{*}{DP}
 & CatBoost            & 0.7595 & 0.7445 & 0.8850 & 0.3130 & \textbf{0.3693} & \textbf{0.3450} \\
 & DT       & 0.7073 & 0.6563 & 0.8537 & 0.3111 & 0.3449 & 0.2853 \\
 & RF       & 0.6850 & 0.6696 & 0.9024 & 0.3203 & 0.3920 & 0.3227 \\
 & LR & 0.7056 & 0.6410 & 0.9024 & \textbf{0.4744} & 0.3397 & 0.2507 \\
 & SVM                 & \textbf{0.8571} & \textbf{0.8129} & 0.9024 & \textbf{0.4744} & \textbf{0.3693} & 0.2408 \\
 & CNN                 & 0.7613 & 0.6895 & 0.8780 & 0.3606 & 0.3658 & 0.3128 \\
 & MLP                 & 0.7561 & 0.6405 & 0.8536 & 0.2381 & 0.2665 & 0.2193 \\
 & Ensemble    & 0.7997 & 0.7575 & \textbf{0.9059} & 0.3740 & \textbf{0.3693} & 0.3000 \\
\hline
\multirow{8}{*}{GA}
 & CatBoost            & 0.9756 & 0.9482 & \textbf{0.9721} & \textbf{0.2465} & 0.9895 & 0.9321 \\
 & DT       & 0.9756 & 0.9482 & 0.8798 & 0.2140 & \textbf{0.9930} & \textbf{0.9565} \\
 & RF       & 0.7317 & 0.6192 & \textbf{0.9721} & \textbf{0.2465} & 0.9895 & 0.9258 \\
 & LR & 0.8048 & 0.6688 & 0.9477 & 0.1946 & 0.9617 & 0.4902 \\
 & SVM                 & \textbf{1.0000} & \textbf{1.0000} & 0.9495 & 0.2435 & 0.9617 & 0.4902 \\
 & CNN                 & 0.9390 & 0.9001 & 0.8954 & 0.2082 & 0.9617 & 0.4902 \\
 & MLP                 & 0.9512 & 0.9453 & 0.9425 & 0.1942 & 0.9617 & 0.4902 \\
 & Ensemble    & 0.9756 & 0.9482 & \textbf{0.9721} & \textbf{0.2465} & 0.9860 & 0.9055 \\
\hline
\multirow{8}{*}{Greedy}
 & CatBoost            & 0.9425 & 0.9450 & 0.8293 & 0.3125 & –      & –      \\
 & DT       & \textbf{0.9477} & \textbf{0.9497} & \textbf{0.8780} & \textbf{0.5027} & –      & –      \\
 & RF       & 0.8014 & 0.6720 & 0.8049 & 0.1859 & –      & –      \\
 & LR & 0.8571 & 0.7036 & 0.8537 & 0.2333 & –      & –      \\
 & SVM                 & \textbf{0.9477} & \textbf{0.9497} & 0.8293 & 0.3549 & –      & –      \\
 & CNN                 & 0.8868 & 0.7446 & 0.8170 & 0.3265 & –      & –      \\
 & MLP                 & 0.8711 & 0.8110 & 0.8240 & 0.2594 & –      & –      \\
 & Ensemble    & \textbf{0.9477} & \textbf{0.9497} & 0.8537 & 0.2689 & –      & –      \\
\hline
\multirow{8}{*}{BnB}
 & CatBoost            & 0.8148 & 0.6612 & 0.8741 & 0.2365 & 0.8407 & 0.6420 \\
 & DT       & 0.6926 & 0.4138 & 0.8963 & 0.2464 & 0.8426 & \textbf{0.6452} \\
 & RF       & 0.7666 & 0.6329 & 0.8741 & 0.2365 & \textbf{0.8481} & 0.6433 \\
 & LR & 0.8333 & 0.6568 & \textbf{0.9222} & \textbf{0.3198} & 0.5944 & 0.3303 \\
 & SVM                 & 0.8333 & 0.6569 & 0.8444 & 0.2322 & 0.6722 & 0.4746 \\
 & CNN                 & 0.7814 & 0.4711 & 0.8241 & 0.2291 & 0.7648 & 0.5736 \\
 & MLP                 & 0.7910 & 0.4718 & 0.8185 & 0.2333 & 0.7611 & 0.5765 \\
 & Ensemble    & \textbf{0.8630} & \textbf{0.7868} & 0.8963 & 0.2396 & 0.4100 & 0.2176 \\
\hline
\multirow{8}{*}{Gurobi}
 & CatBoost            & \textbf{1.0000} & \textbf{1.0000} & –      & –      & \textbf{0.4591} & 0.3925 \\
 & DT       & 0.9961 & 0.4990 & –      & –      & 0.4474 & 0.4247 \\
 & RF       & \textbf{1.0000} & \textbf{1.0000} & –      & –      & \textbf{0.4591} & \textbf{0.4314} \\
 & LR & \textbf{1.0000} & \textbf{1.0000} & –      & –      & 0.4339 & 0.3220 \\
 & SVM                 & \textbf{1.0000} & \textbf{1.0000} & –      & –      & 0.4339 & 0.3459 \\
 & CNN                 & \textbf{1.0000} & \textbf{1.0000} & –      & –      & 0.4144 & 0.3829 \\
 & MLP                 & \textbf{1.0000} & \textbf{1.0000} & –      & –      & 0.4105 & 0.3928 \\
 & Ensemble    & \textbf{1.0000} & \textbf{1.0000} & –      & –      & 0.4514 & \textbf{0.4314} \\
\hline
\multirow{8}{*}{OR-Tools}
 & CatBoost            & 0.9500 & 0.4871 & \textbf{0.9250} & 0.2403 & 0.7268 & 0.3459 \\
 & DT       & 0.9500 & 0.3263 & 0.8500 & 0.1889 & 0.6912 & 0.3695 \\
 & RF       & 0.9500 & 0.4872 & \textbf{0.9250} & 0.2403 & \textbf{0.7786} & \textbf{0.5042} \\
 & LR & 0.9500 & 0.4872 & \textbf{0.9250} & 0.2403 & 0.6946 & 0.2862 \\
 & SVM                 & 0.9500 & 0.4872 & 0.8750 & 0.2365 & 0.6750 & 0.2979 \\
 & CNN                 & 0.9500 & 0.4872 & 0.9089 & \textbf{0.3695} & 0.6607 & 0.2870 \\
 & MLP                 & \textbf{0.9750} & \textbf{0.8268} & 0.9000 & 0.2400 & 0.6714 & 0.3025 \\
 & Ensemble    & 0.9500 & 0.4872 & \textbf{0.9250} & 0.2403 & 0.7339 & 0.4258 \\
\hline
\end{tabular}
}
\label{tab:all_models_per_solver_classification}
\end{table}
In both profit-maximization and minimization variants, top-3 ensemble classifiers demonstrate consistent and robust performance across metrics. In the maximization variant, ensembles like MLP-RF-CatBoost and CatBoost-SVM-LR achieve top F1-scores for $O_s$ in GA (0.2465) and $T_s$ in BnB (0.7868), respectively. In $T_s$ prediction for Gurobi, imbalanced class distribution leads most models to report perfect F1-scores. However, classification fails for $M_s$ in greedy and $O_s$ in Gurobi due to constant target values. While some standalone models perform well in specific cases, they lack consistency. For instance, SVM consistently performs well in predicting all the metrics for DP. DP has a pseudo-polynomial continuous time-complexity pattern for which models like SVM, which are efficient in handling non-linear patterns, perform well. Similarly, Catboost, DT, and RF show strong performance in specific cases but lack consistency across all the algorithms. DT and SVMs show effectiveness in high-dimensional spaces (e.g., for $T_s$), but struggle when classes are not fully linearly separable (e.g., for $O_s$).

Similar trends are observed in the minimization variant. For DP, the top-3 ensemble CNN-MLP-Catboost achieve an F1-score of 0.6751, increasing the average F1-score by a minimum of 0.2. Ensembles of top-3 classifiers consistently perform well in the case of predicting $T_s$ and $O_s$ for most of the algorithms with F1-scores above 0.5 ($T_s$) and 0.5 ($M_s$), respectively. Although in this variant, many standalone classifiers also perform equally well for some algorithms (particularly in predicting $T_s$), the results are not consistent across all the metrics. Furthermore, depending on the algorithm's behavior, the ensemble of top-3 models yields a perfect score in many cases. For instance, models achieve a perfect F1-score in $O_s$ prediction for DP and OR-Tools due to constant target values. In other cases, such as $O_s$ of GA, models skip classification since only a single class exists. Furthermore, in $M_s$ of GA and $T_s$ of OR-Tools predictions, performance metrics are tightly clustered around a single class, leading the models to overfit. Results are shown in Table \ref{tab:all_models_per_solver_classification} and \ref{tab:combined_minimization_all_models}.

\subsection{Regression vs RL-based Model Performance}

\begin{table}
\caption{Models performance for each algorithm (Regression $|$ Maximization)}
\centering
\scriptsize
\setlength{\tabcolsep}{2pt}{
\begin{tabular}{|c|l|cc|cc|cc|}
\hline
\textbf{Algorithm} & \textbf{Model} & \multicolumn{2}{c|}{\textbf{Solution Time}} & \multicolumn{2}{c|}{\textbf{Optimality Gap}} & \multicolumn{2}{c|}{\textbf{Peak Memory}} \\
\cline{3-8}
 & & \textbf{RMSE} & $\mathbf{R^2}$ & \textbf{RMSE} & $\mathbf{R^2}$ & \textbf{RMSE} & $\mathbf{R^2}$ \\
\hline
\multirow{10}{*}{DP} & CatBoost & 149.2524 & 0.9774  & \textbf{0.0001} & -0.0428 & \textbf{362548.1811} & \textbf{-0.1448} \\
& DT & 209.3437 & 0.9555 & 0.0002 & -1.8362 & 553143.5670 & -1.6649 \\
& RF & 163.2385 & 0.9730 & \textbf{0.0001} & -0.7205 & 374605.2160 & -0.2222 \\
& LR & 1767.2511 & -2.1673 & 0.0002 & -3.1359 & 426954.5486 & -0.5877 \\
& SVM & 217.9979 & 0.9518 & \textbf{0.0001} & -0.0965 & 484541.9661 & -1.0449 \\
& CNN & 162.9699 & 0.9731 & \textbf{0.0001} & -0.6854 & 452723.9149 & -0.7852 \\
& MLP & 414.7251 & 0.8256 & \textbf{0.0001} & \textbf{0.4139} & 537542.0154 & -1.5167\\
& SARSA & 1409.5355 & -1.0149 & \textbf{0.0001} & -0.6212 & 498127.6508 & -1.1612 \\
& Q-Learning & 1409.5355 & -1.0149 & \textbf{0.0001} & -0.6212 & 498127.6508 & -1.1612 \\
& Ensembled  & \textbf{132.4580}          & \textbf{0.9822  }        & \textbf{0.0001}       & -0.4336      & 401047.4089  & -0.4009 \\
\hline
\multirow{10}{*}{GA} & CatBoost & 0.3426 & 0.9866 & 2.3626 & -0.2298 & 171.9897 & 0.8087 \\
& DT & 0.2584 & 0.9924 & 3.5303 & -1.7458 & 243.9730 & 0.6150 \\
& RF & \textbf{0.1782} & \textbf{0.9964} & 2.6188 & -0.5110 & \textbf{170.8677} & \textbf{0.8111} \\
& LR & 0.5483 & 0.9656 & 4.1256 & -2.7499 & 377.7469 & 0.0770 \\
& SVM & 0.9147 & 0.9044 & 2.7109 & -0.6192 & 394.3134 & -0.0057 \\
& CNN & 0.3410 & 0.9867 & 2.8731 & -0.8187 & 400.9455 & -0.0399 \\
& MLP & 1.0059 & 0.8843 & 3.4839 & -1.6741 & 400.9399 & -0.0398 \\
& SARSA & 4.2248 & -1.0403 & 2.5678 & -0.4527 & 1807.9769 & -20.1440 \\
& Q-Learning & 4.2248 & -1.0403 & 2.5678 & -0.4527 & 1807.9769 & -20.1440 \\
 & Ensembled  & 0.5294            & 0.9680          & \textbf{2.3557}      & \textbf{-0.2227}      & 291.3104     & 0.4511 \\
\hline
\multirow{10}{*}{Greedy} & CatBoost & 0.0104 & 0.9798 & \textbf{2.2861} & \textbf{0.1451} & - & - \\
& DT & 0.0074 & 0.9896 & 2.8466 & -0.3255 &  - & - \\
& RF & 0.0065 & 0.9920 & 2.3140 & 0.1241 &  - & - \\
& LR & \textbf{0.0059} & \textbf{0.9935} & 2.7568 & -0.2432 &  - & - \\
& SVM & 0.0064 & 0.9924 & 2.4355 & 0.0297 &  - & - \\
& CNN & 0.0099 & 0.9814 & 2.6266 & -0.1286 &  - & - \\
& MLP & 0.0218 & 0.9106 & 3.2778 & -0.7576 &  - & - \\
& SARSA & 0.1262 & -2.0010 & 2.4772 & -0.0039 &  - & - \\
& Q-Learning & 0.1262 & -2.0010 & 2.4772 & -0.0039 &  - & - \\
& Ensembled  & 0.0110            & 0.9772       & 2.4025     & 0.0558       &  - & -  \\
\hline
\multirow{10}{*}{BnB} & CatBoost & \textbf{27.0357} & \textbf{0.5999} & 1.6184 & 0.2709 & \textbf{724.5794} & \textbf{0.3059} \\
& DT & 49.3764 & -0.3344 & 1.4117 & 0.4452 & 810.1949 & 0.1322 \\
& RF & 28.3552 & 0.5599 & \textbf{0.9782} & \textbf{0.7336} & 729.1758 & 0.2971 \\
& LR & 36.5750 & 0.2678 & 2.3427 & -0.5278 & 830.7786 & 0.0875 \\
& SVM & 32.1808 & 0.4332 & 1.8615 & 0.0353 & 807.3278 & 0.1383 \\
& CNN & 27.8583 & 0.5752 & 2.2379 & -0.3942 & 770.0387 & 0.2161 \\
& MLP & 41.8760 & 0.0402 & 2.4631 & -0.6890 & 769.6135 & 0.2169 \\
& SARSA & 42.7719 & -0.0013 & 1.9477 & -0.0561 & 1261.7529 & -1.1047 \\
& Q-Learning & 42.7719 & -0.0013 & 1.9477 & -0.0561 & 1261.7529 & -1.1047 \\
& Ensembled  & 28.3155           & 0.5612          & 1.9179       & -0.0240      & 835.4052     & 0.0773  \\

\hline
\multirow{10}{*}{Gurobi} & CatBoost & 0.0231 & 0.1771 & - & - & \textbf{16710.3450} & \textbf{0.4444}\\
& DT & 0.0260 & -0.1423 &  - & - & 22658.9210 & -0.215 \\
& RF & 0.0306 & -0.4406 &  - & - & 17545.7276 & 0.3875 \\
& LR & 0.0264 & -0.0711 &  - & - & 20523.5531 & 0.1620 \\
& SVM & 0.0213 & 0.3026 &  - & - & 19087.2451 & 0.2752 \\
& CNN & 0.0229 & 0.1936 &  - & - & 19439.0274 & 0.2482 \\
& MLP & 0.0280 & -0.2081 &  - & - & 18565.4146 & 0.3142 \\
& SARSA & 0.0544 & -3.5485 &  - & - & 49959.1176 & -3.9658 \\
& Q-Learning & 0.0544 & -3.5485 &  - & - & 49959.1176 & -3.9658 \\
 & Ensembled  & \textbf{0.0211}           & \textbf{0.3133 }         &  - & -       & 20188.2146   & 0.1891  \\
\hline
\multirow{10}{*}{OR-Tools} & CatBoost & 40.3333 & 0.1348 & 0.0016 & 0.0101 & \textbf{102358.9957} & \textbf{0.2778} \\
& DT & 43.0383 & 0.0148 & \textbf{0.0000} & \textbf{1.0000} & 125716.5600 & 0.0148 \\
& RF & 39.8168 & 0.1568 & 0.0017 & -0.1483 & 114190.7748 & 0.1012 \\
& LR & 40.4287 & 0.1307 & 0.0023 & -1.0741 & 132905.6234 & -0.2176 \\
& SVM & 41.8249 & 0.0696 & 0.0013 & 0.3066 & 122018.0982 & -0.0263 \\
& CNN & 44.1727 & -0.0378 & 0.0009 & 0.6412 & 122730.4497 & -0.0383 \\
& MLP & 38.2893 & 0.2203 & 0.0010 & 0.6254 & 122526.8673 & -0.0349 \\
& SARSA & 43.4489 & -0.0040 & 0.0016 & -0.0654 & 241063.7621 & -3.0057 \\
& Q-Learning & 43.4489 & -0.0040 & 0.0016 & -0.0654 & 241063.7621 & -3.0057 \\
 & Ensembled  & \textbf{37.6704}          & \textbf{0.2452}         & 0.002 & 0.1845 & 113879.3239 & 0.1061\\
\hline
\end{tabular}
}
\label{tab:all_models_per_solver_regression}
\end{table}

\begin{table*}[t]
  \caption{Ensemble model performance for each algorithm (Classification \& Regression $|$ Minimization)}
  \centering

  \resizebox{\textwidth}{!}{%
    \begin{tabular}{|l|cccc|cccc|cccc|}
      \hline
      \multirow{2}{*}{\textbf{Algorithm}}
        & \multicolumn{4}{c|}{\textbf{Solution Time}}
        & \multicolumn{4}{c|}{\textbf{Optimality Gap}}
        & \multicolumn{4}{c|}{\textbf{Peak Memory}} \\
      \cline{2-13}
      & \textbf{Acc} & \textbf{F1-Score} & \textbf{RMSE} & $\mathbf{R^2}$
      & \textbf{Acc} & \textbf{F1-Score} & \textbf{RMSE} & $\mathbf{R^2}$
      & \textbf{Acc} & \textbf{F1-Score} & \textbf{RMSE} & $\mathbf{R^2}$ \\
      \hline
      DP
        & 0.8232 & 0.6751 & 0.8265       & 0.9494
        & 1.0000 & 1.0000 & 0.0051       & -313.6447
        & 0.9286 & 0.5555 & 184.4943     & 0.8521 \\
      \hline
      GA  
        & 0.9875 & 0.9808 & 0.0366       & 0.9981
        & - & - & 20704530.0     & -16.1434
        & 1.0000 & 1.0000 & 339.4771     & 0.5857 \\
      \hline
      Greedy             
        & 0.9964 & 0.9915 & 0.0040       & 0.9959
        & 0.7000 & 0.3333 & 24.5256      & -0.2562  
        & –      & –      & -       & - \\
      \hline
      BnB  
        & 0.9839 & 0.9671 & 0.0000       & 0.9927
        & 0.9250 & 0.7606 & 8.6131       & 0.8953   
        & –      & –      & -      & - \\
      \hline
      Gurobi             
        & 0.8000 & 0.4797 & 0.0246       & 0.5008
        & –      & –      & -       & -   
        & 0.2482 & 0.2200 & 21164.92     & 0.0779 \\
      \hline
      OR-Tools     
        & 1.0000 & 1.0000 & 12.7617      & -41.4733
        & 1.0000 & 1.0000 & 0.0076       & -12.7034
        & 0.9357 & 0.5526 & 42527.64     & 0.0722 \\
      \hline
    \end{tabular}%
}
  \label{tab:combined_minimization_all_models}
\end{table*}

In the regression-based framework, top-3 ensemble regressors outperform standalone models in both 0-1 KP variants by delivering more stable predictions. In the maximization variant, ensembles significantly reduce RMSE (e.g., 132.4580 for $T_s$ in DP, 2.3557 for $O_s$ in GA) while maintaining high $R^2$ values. In the case of DP, the RMSE for $O_s$ is extremely low (0.0001) since the true values are nearly constant (close to 0 due to scaling), making the small errors insignificant. Meanwhile, standalone regressors show limitations. LR performs poorly on algorithms with non-linear patterns (e.g., predicting $O_s$), resulting in negative $R^2$ values. RL methods (Q-learning, SARSA) underperform, with negative $R^2$ across all tasks as shown in Table \ref{tab:all_models_per_solver_regression}. However, CatBoost performs well for $M_s$ due to its handling of categorical features.

In the minimization variant, ensembles also excel, especially for $T_s$ and $O_s$. In the case of greedy and BnB, ensembles with RMSE (0.0040, 0.000007) perform equally well compared to standalone regressors like SVM and LR, with slightly higher $R^2$ values, confirming the robustness of ensembles on very low-error tasks. In the case of $O_s$ prediction for GA and BnB, ensembles exceed the standalone regressors with low RMSE values; however, $R^2$ remains negative due to outliers. Thus, for both the variants, consistency makes ensembles well-suited compared to standalone regressors due to their ability to handle bias or minimal variance.

\section{Ablation Study}
\label{sec_ablation_study}
\subsection{Effect of Hyperparameter Tuning on the Models}
We tune each base classifier using grid search, focusing on key hyperparameters specific to each model (e.g., max\_depth, n\_estimators and n\_sample\_split for RF/DT, penalty and kernel for SVM, epochs (50 and 100) for MLP/CNN, regularization strength for LR, and learning rate, depth and l2\_leaf\_reg for Catboost). In the maximization variant, RF and CatBoost improve their F1-scores for $M_s$ in Gurobi by 0.4, while DT improves by 0.4 for $T_s$ in DP. Increasing MLP/CNN epochs from 50 to 100 boosts F1-scores by at least 0.2 on average. We observe similar improvements in the models trained for the minimization variant as well. After tuning the hyperparameters, the F1 Score for each of the models improves by a minimum of 0.1 across all the algorithms for both variants. These tuned models are used in the final ensembles.
\subsection{Impact of Ensemble Techniques on the Overall Framework}
We evaluate ensemble performance using top-\{3, 5, 7\} models for both classifiers and regressors. Top-3 ensembles consistently deliver the most stable results across both 0-1 KP variants. For example, in the profit-maximization variant, top-3 classifiers achieve an F1-score of 0.7868 for BnB, outperforming both standalone and larger ensembles. Larger ensembles underperform as they include models that do not consistently perform well across tasks. For instance, in case of $M_s$ prediction for OR-Tools, top-3 ensembles achieve an F1-score of 0.42, followed by top-5 ensembles with 0.35 then top-7 ensembles with 0.27 F1-scores, respectively. Similarly, top-3 regressors show strong RMSE and $R^2$ performance across metrics and outperform larger ensembles in consistency. For fair comparison, results in Table \ref{tab:all_models_per_solver_classification} and \ref{tab:all_models_per_solver_regression} report the top-3 ensemble models.

\subsection{Feature Interpretation across Performance Metrics}
\begin{figure*}[t]
  \centering
  % first subfigure
  \begin{subfigure}[b]{0.33\textwidth}
    \includegraphics[width=\linewidth]{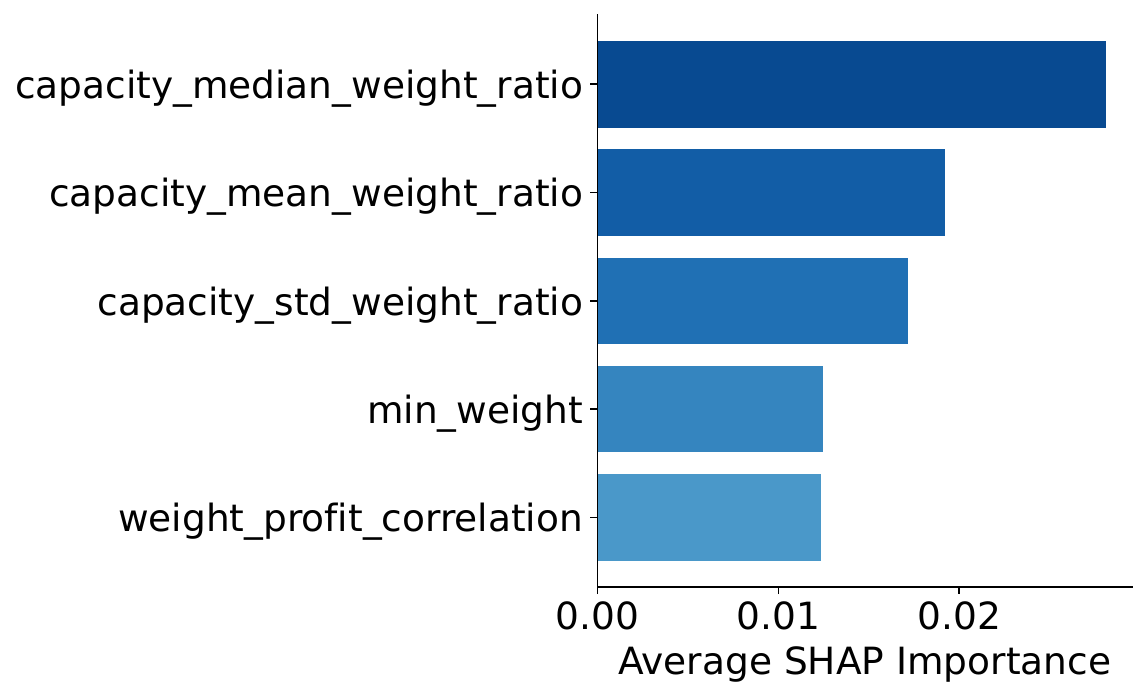}
    \caption{Optimality Gap}
    \label{fig:opt-gap}
  \end{subfigure}%
  \hfill
  % second subfigure
  \begin{subfigure}[b]{0.33\textwidth}
    \includegraphics[width=\linewidth]{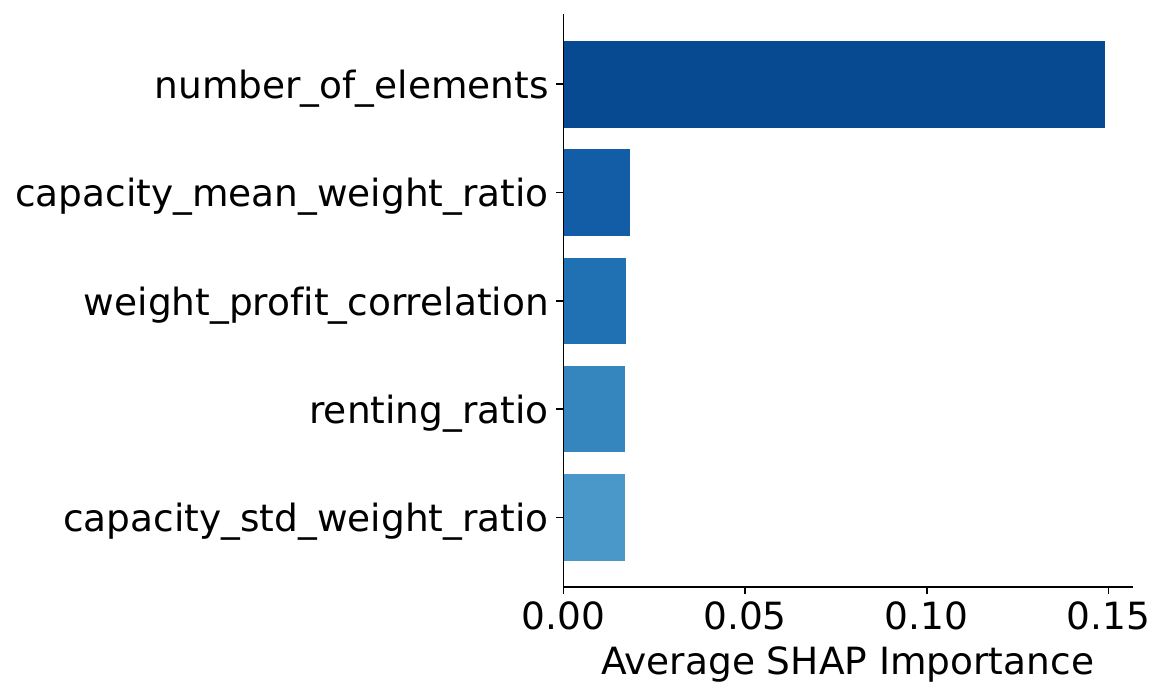}
    \caption{Solution Time}
    \label{fig:sol-time}
  \end{subfigure}%
  \hfill
  % third subfigure
  \begin{subfigure}[b]{0.33\textwidth}
    \includegraphics[width=\linewidth]{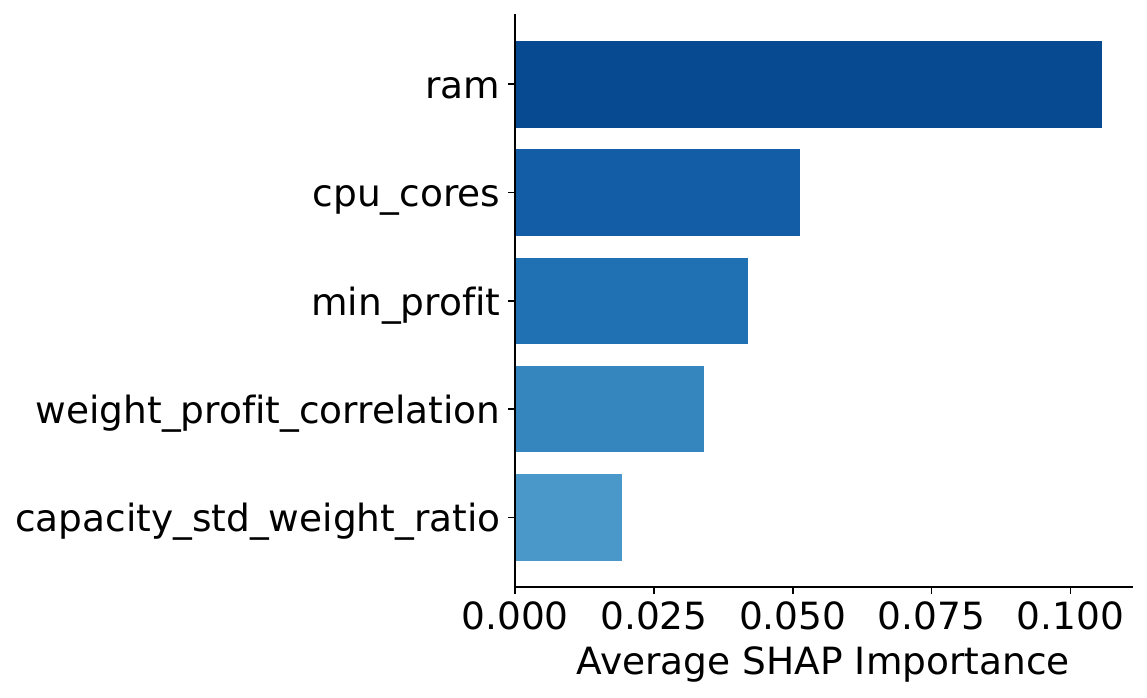}
    \caption{Peak Memory}
    \label{fig:peak-mem}
  \end{subfigure}
  \caption{Top‐5 SHAP-based features for each metric across all models and algorithms}
  \label{fig:feature_importance}
\end{figure*}
For explainability, we use SHAP \cite{NIPS2017_8a20a862} analysis on top-performing ensemble classifiers. Figure \ref{fig:feature_importance} shows the list of top-5 features for each of the performance metrics across all the models and algorithms. In the case of predicting the $O_s$, our experiments show that features related to capacity-weight ratios and weight-profit correlation contribute highly to the model's accuracy. With a high weight-profit correlation, heuristic-based algorithms tend to pick near-optimal sets faster, whereas a low correlation forces them to explore more solutions, impacting $T_s$ and $M_s$ predictions. Similarly, capacity-weight ratios, instance-specific features like the number of elements (directly impacting the runtime of an algorithm), and renting ratio influence the prediction of $T_s$ by defining the complexity levels and size of the search space for the heuristic-based algorithms. For example, heuristic-based algorithms tend to converge faster in the case of instances with a high renting ratio \cite{huerta2022improving}, reducing runtime. Notably, for $M_s$ prediction, hardware-specific features, like RAM and CPU cores, appear to be the most impactful, as exploration of larger tables and execution of parallel data structures for algorithms like GA, Gurobi highly influences the memory usage.

\subsection{Zero-shot Inference Performance from LLM}
\begin{table*} [t]
  \caption{Performance evaluation of LLMs (Regression $|$ Maximization)}
  \centering
  \scriptsize
% \resizebox{\textwidth}{!}{
    \begin{tabular}{|l|ccc|ccc|ccc|}
      \hline
      \multirow{2}{*}{\textbf{Algorithm}}
        & \multicolumn{3}{c|}{\textbf{Solution Time}}
        & \multicolumn{3}{c|}{\textbf{Optimality Gap}}
        & \multicolumn{3}{c|}{\textbf{Peak Memory}} \\
      \cline{2-10}
      & \textbf{GPT-4} & \textbf{Gemini} & \textbf{Regressor}
      & \textbf{GPT-4} & \textbf{Gemini} & \textbf{Regressor}
      & \textbf{GPT-4} & \textbf{Gemini} & \textbf{Regressor} \\
      \hline
      DP
        & 375614.33 & 220439.41 & 151.93
        & 0.00 & 0.76 & 0.00
        & 965280.31 & 29963608.36 & 381250.69\\
        \hline
      GA  
        & 12175.74 & 255885.91 & 4.35
        & 4.48 & 4.49 & 3.38
        & 688754.85 & 24068.73 & 12250.24\\
        \hline
      Greedy             
        & 59.58 & 5.56 & 0.01
        & 3.22 & 2.72 & 2.84
        & 3089.87 & 4036.95 & -\\
        \hline
      BnB   
        & 21189.00 & 199862.23 & 24.90
        & 1.83 & 1.85 & 1.38
        & 688299.49 & 490037.11 & 1026.10\\
        \hline
      Gurobi             
        & 6157.43 & 90098.70 & 0.03
        & 0.00 & 0.00 & -
        & 471668.71 & 140818.29 & 19057.90\\
        \hline
      OR-Tools     
        & 10030.75 & 135623.65 & 37.38
        & 0.00 & 0.00 & 0.51
        & 549512.87 & 320023.78 & 112542.79\\
      \hline
    \end{tabular}% 
    
  \label{tab:llm_results}
\end{table*}
Given the rapid advancement of LLMs, we evaluated GPT-4o and Gemini-2.5-Flash for predicting $T_s$, $O_s$, and $M_s$ using zero-shot inference. Using an independent prediction approach, both LLMs achieved performance close to the best-trained regressors for $O_s$ prediction. Notably, GPT-4o accurately predicted $O_s$ for DP, which typically yields optimal solutions, while Gemini did not. However, both models showed significantly higher errors in predicting $T_s$ and $M_s$ compared to trained regressors, as shown in Table \ref{tab:llm_results}.

\section{Applicability of Our Framework}
\label{sec_applicability_framework}
 To demonstrate the framework's applicability, we generated predictions for 10 random instances using fixed hardware settings (CPU cores = 8 and RAM = 128). Though the full SLA negotiation process is not implemented in the current version, we assume static SLA thresholds to simulate predefined agreements between the CSP and the client. The thresholds for solving the set of instances are \(T_i^{\max} = 100\text{ s}, O_i^{\max} = 3.5\%, M_i^{\max} = 20000\text{ KB}.\) Table \ref{tab:framework_output} shows the sample output generated by our predictive model for an instance. 

While our framework currently focuses on generating performance predictions, we suggest possible decision strategies, such as rule-based filtering or best-fit ranking for the `Decider' module. In the sample output (Table \ref{tab:framework_output}), both BnB and GA meet SLA requirements, allowing CSPs to select the most cost-effective option to finalize SLA-compliant algorithm execution.
 \begin{table}[htbp]
\caption{Predicted performance and SLA compliance for an instance under given SLA thresholds}
\centering
\resizebox{\columnwidth}{!}{
\begin{tabular}{|l|l|c|c|}
\hline
\textbf{Algorithm} & \textbf{Metric} & \textbf{Predicted Value} & \textbf{SLA Met?} \\
\hline
\multirow{3}{*}{Greedy} 
  & Solution Time     & 0.23                  & Yes \\
  & Optimality Gap  & 1.60                  & Yes \\
  & Peak Memory      & -       & No  \\
\hline
\multirow{3}{*}{DP}    
  & Solution Time     & 373.38                & No  \\
  & Optimality Gap  & 0.00                  & Yes \\
  & Peak Memory      & 1375027.2         & No  \\
\hline
\multirow{3}{*}{BnB}    
  & Solution Time     & 29.46                 & Yes \\
  & Optimality Gap   & 1.13                  & Yes \\
  & Peak Memory     & 11424.0               & Yes \\
\hline
\multirow{3}{*}{Gurobi} 
  & Solution Time     & 0.10                 & Yes \\
  & Optimality Gap   & 0.00                  & Yes \\
  & Peak Memory     & 179146.0               & No \\
\hline
\multirow{3}{*}{OR-Tools}      
  & Solution Time     & 30.00                 & Yes \\
  & Optimality Gap   & 0.00                & Yes \\
  & Peak Memory      & 259685.2               & No \\
\hline
\multirow{3}{*}{GA}      
  & Solution Time    & 1.14                & Yes  \\
  & Optimality Gap   & 3.14                  & Yes  \\
  & Peak Memory     & 12083.2              & Yes \\
\hline
\end{tabular}
}
\label{tab:framework_output}
\end{table}

\section{Conclusion and Future Work}
\label{sec_conclusions}
This study presents an SLA-centric framework for automated algorithm selection in cloud environments. By predicting key performance metrics and leveraging an SLA-aware decision module, the framework enables efficient resource-aware algorithm selection. Experimental results show that ensemble classifiers consistently outperform standalone models. While our ML-based algorithm selection framework shows strong potential, it has limitations. It relies on fixed hardware settings, which may not reflect the dynamic nature of real-world cloud environments. Additionally, assuming static resource availability may impact SLA compliance in distributed, shared settings.

As future work, we suggest evaluating the framework on real-world datasets with dynamic resource conditions and incorporating real-time traces while performing SLA negotiations. We also propose to investigate the viability of LLM-based predictors for SLA-driven decisions and improve ensemble regression strategies for more robust and efficient inference.

\bibliographystyle{IEEEtran.bst}

\end{document}